\documentclass{article}

\usepackage{amsmath}
\usepackage{float}
\usepackage{placeins}
\usepackage{verbatim}
\usepackage{graphicx}
\usepackage{multirow}

\usepackage{microtype}
\usepackage{graphicx}
\usepackage{xcolor}
\usepackage{subfigure}
\usepackage{booktabs} 

\usepackage{hyperref}

\usepackage[arxiv]{icml2019}

\icmltitlerunning{The State of Sparsity in Deep Neural Networks}

\begin{document}

\twocolumn[
\icmltitle{The State of Sparsity in Deep Neural Networks}



\icmlsetsymbol{equal}{*}
\icmlsetsymbol{resident}{$\dagger$}

\begin{icmlauthorlist}
\icmlauthor{Trevor Gale}{equal,gbrain,resident}
\icmlauthor{Erich Elsen}{equal,deepmind}
\icmlauthor{Sara Hooker}{gbrain,resident}
\end{icmlauthorlist}

\icmlaffiliation{gbrain}{Google Brain}
\icmlaffiliation{deepmind}{DeepMind}

\icmlcorrespondingauthor{Trevor Gale}{tgale@google.com}

\icmlkeywords{Neural networks, sparsity}

\vskip 0.3in
]



\printAffiliationsAndNotice{\icmlEqualContribution\airesident} 

\begin{abstract}

We rigorously evaluate three state-of-the-art techniques for inducing sparsity in deep neural networks on two large-scale learning tasks: Transformer trained on WMT 2014 English-to-German, and ResNet-50 trained on ImageNet. Across thousands of experiments, we demonstrate that complex techniques \cite{variational-dropout, l0-regularization} shown to yield high compression rates on smaller datasets perform inconsistently, and that simple magnitude pruning approaches achieve comparable or better results. Based on insights from our experiments, we achieve a new state-of-the-art sparsity-accuracy trade-off for ResNet-50 using only magnitude pruning. Additionally, we repeat the experiments performed by \citet{lottery-ticket-hypothesis} and \citet{rethinking-pruning} at scale and show that unstructured sparse architectures learned through pruning cannot be trained from scratch to the same test set performance as a model trained with joint sparsification and optimization. Together, these results highlight the need for large-scale benchmarks in the field of model compression. We open-source our code, top performing model checkpoints, and results of all hyperparameter configurations to establish rigorous baselines for future work on compression and sparsification.
 
\end{abstract}

\setcounter{footnote}{1}

\section{Introduction}

Deep neural networks achieve state-of-the-art performance in a variety of domains including image classification \cite{rn50}, machine translation \cite{transformer}, and text-to-speech \cite{wavenet, wavernn}. While model quality has been shown to scale with model and dataset size \cite{deep-learning-scaling}, the resources required to train and deploy large neural networks can be prohibitive. State-of-the-art models for tasks like image classification and machine translation commonly have tens of millions of parameters, and require billions of floating-point operations to make a prediction for a single input sample.

Sparsity has emerged as a leading approach to address these challenges. By sparsity, we refer to the property that a subset of the model parameters have a value of exactly zero\footnote{The term sparsity is also commonly used to refer to the proportion of a neural network’s weights that are zero valued. Higher sparsity corresponds to fewer weights, and smaller computational and storage requirements. We use the term in this way throughout this paper.}. With zero valued weights, any multiplications (which dominate neural network computation) can be skipped, and models can be stored and transmitted compactly using sparse matrix formats. It has been shown empirically that deep neural networks can tolerate high levels of sparsity \cite{lwac, exploring-sparsity-rnn, sws}, and this property has been leveraged to significantly reduce the cost associated with the deployment of deep neural networks, and to enable the deployment of state-of-the-art models in severely resource constrained environments \cite{fisher-pruning, wavernn, lpcnet}.

Over the past few years, numerous techniques for inducing sparsity have been proposed and the set of models and datasets used as benchmarks has grown too large to reasonably expect new approaches to explore them all. In addition to the lack of standardization in modeling tasks, the distribution of benchmarks tends to slant heavily towards convolutional architectures and computer vision tasks, and the tasks used to evaluate new techniques are frequently not representative of the scale and complexity of real-world tasks where model compression is most useful. These characteristics make it difficult to come away from the sparsity literature with a clear understanding of the relative merits of different approaches.

In addition to practical concerns around comparing techniques, multiple independent studies have recently proposed that the value of sparsification in neural networks has been misunderstood \cite{lottery-ticket-hypothesis, rethinking-pruning}. While both papers suggest that sparsification can be viewed as a form of neural architecture search, they disagree on what is necessary to achieve this. Specifically, \citet{rethinking-pruning} re-train learned sparse topologies with a random weight initialization, whereas \citet{lottery-ticket-hypothesis} posit that the exact random weight initialization used when the sparse architecture was learned is needed to match the test set performance of the model sparsified during optimization.

In this paper, we address these ambiguities to provide a strong foundation for future work on sparsity in neural networks. \textbf{Our main contributions:} (1) We perform a comprehensive evaluation of variational dropout \cite{variational-dropout}, $l_0$ regularization \cite{l0-regularization}, and magnitude pruning \cite{to-prune-or-not} on Transformer trained on WMT 2014 English-to-German and ResNet-50 trained on ImageNet. To the best of our knowledge, we are the first to apply variational dropout and $l_0$ regularization to models of this scale. While variational dropout and $l_0$ regularization achieve state-of-the-art results on small datasets, we show that they perform inconsistently for large-scale tasks and that simple magnitude pruning can achieve comparable or better results for a reduced computational budget. (2) Through insights gained from our experiments, we achieve a new state-of-the-art sparsity-accuracy trade-off for ResNet-50 using only magnitude pruning. (3) We repeat the lottery ticket \cite{lottery-ticket-hypothesis} and scratch \cite{rethinking-pruning} experiments on Transformer and ResNet-50 across a full range of sparsity levels. We show that unstructured sparse architectures learned through pruning cannot be trained from scratch to the same test set performance as a model trained with pruning as part of the optimization process. (4) We open-source our code, model checkpoints, and results of all hyperparameter settings to establish rigorous baselines for future work on model compression and sparsification \footnote{\url{https://bit.ly/2ExE8Yj}}.

\section{Sparsity in Neural Networks}
We briefly provide a non-exhaustive review of proposed approaches for inducing sparsity in deep neural networks.

Simple heuristics based on removing small magnitude weights have demonstrated high compression rates with minimal accuracy loss \cite{sparse-connection-1997, memory-bounded-convnet, lwac}, and further refinement of the sparsification process for magnitude pruning techniques has increased achievable compression rates and greatly reduced computational complexity \cite{dynamic-network-surgery, to-prune-or-not}.

Many techniques grounded in Bayesian statistics and information theory have been proposed \cite{variational-information-bottleneck, variational-dropout, l0-regularization, bayesian-compression, sws}. These methods have achieved high compression rates while providing deep theoretical motivation and connections to classical sparsification and regularization techniques. 

Some of the earliest techniques for sparsifying neural networks make use of second-order approximation of the loss surface to avoid damaging model quality \cite{optimal-brain-damage, optimal-brain-surgeon}. More recent work has achieved comparable compression levels with more computationally efficient first-order loss approximations, and further refinements have related this work to efficient empirical estimates of the Fisher information of the model parameters \cite{pruning-convnet-nvidia, fisher-pruning}. 

Reinforcement learning has also been applied to automatically prune weights and convolutional filters \cite{runtime-neural-pruning, automatic-model-compression}, and a number of techniques have been proposed that draw inspiration from biological phenomena, and derive from evolutionary algorithms and neuromorphic computing \cite{dynamic-network-surgery, deep-rewiring, sparse-evolutionary-training}.

A key feature of a sparsity inducing technique is if and how it imposes structure on the topology of sparse weights. While unstructured weight sparsity provides the most flexibility for the model, it is more difficult to map efficiently to parallel processors and has limited support in deep learning software packages. For these reasons, many techniques focus on removing whole neurons and convolutional filters, or impose block structure on the sparse weights \cite{network-slimming, thinet, blocksparse-gpu-kernels}. While this is practical, there is a trade-off between achievable compression levels for a given model quality and the level of structure imposed on the model weights. In this work, we focus on unstructured sparsity with the expectation that it upper bounds the compression-accuracy trade-off achievable with structured sparsity techniques.

\section{Evaluating Sparsification Techniques at Scale}

As a first step towards addressing the ambiguity in the sparsity literature, we rigorously evaluate magnitude-based pruning \cite{to-prune-or-not}, sparse variational dropout \cite{variational-dropout}, and $l_0$ regularization \cite{l0-regularization} on two large-scale deep learning applications: ImageNet classification with ResNet-50 \cite{rn50}, and neural machine translation (NMT) with the Transformer on the WMT 2014 English-to-German dataset \cite{transformer}. For each model, we also benchmark a random weight pruning technique, representing the lower bound of compression-accuracy trade-off any method should be expected to achieve.

Here we briefly review the four techniques and introduce our experimental framework. We provide a more detailed overview of each technique in Appendix \ref{appendix_a}.

\subsection{Magnitude Pruning}

Magnitude-based weight pruning schemes use the magnitude of each weight as a proxy for its importance to model quality, and remove the least important weights according to some sparsification schedule over the course of training. For our experiments, we use the approach introduced in \citet{to-prune-or-not}, which is conveniently available in the TensorFlow model\_pruning library \footnote{\url{https://bit.ly/2T8hBGn}}. This technique allows for masked weights to reactivate during training based on gradient updates, and makes use of a gradual sparsification schedule with sorting-based weight thresholding to achieve a user specified level of sparsification. These features enable high compression ratios at a reduced computational cost relative to the iterative pruning and re-training approach used by \citet{lwac}, while requiring less hyperparameter tuning relative to the technique proposed by \citet{dynamic-network-surgery}.

\subsection{Variational Dropout}

Variational dropout was originally proposed as a re-interpretation of dropout training as variational inference, providing a Bayesian justification for the use of dropout in neural networks and enabling useful extensions to the standard dropout algorithms like learnable dropout rates \cite{variational-dropout-local-reparameterization}. It was later demonstrated that by learning a model with variational dropout and per-parameter dropout rates, weights with high dropout rates can be removed post-training to produce highly sparse solutions \cite{variational-dropout}. 

Variational dropout performs variational inference to learn the parameters of a fully-factorized Gaussian posterior over the weights under a log-uniform prior. In the standard formulation, we apply a local reparameterization to move the sampled noise from the weights to the activations, and then apply the additive noise reparameterization to further reduce the variance of the gradient estimator. Under this parameterization, we directly optimize the mean and variance of the neural network parameters. After training a model with variational dropout, the weights with the highest learned dropout rates can be removed to produce a sparse model.

\subsection{$l_0$ Regularization}

$l_0$ regularization explicitly penalizes the number of non-zero weights in the model to induce sparsity. However, the $l_0$-norm is both non-convex and non-differentiable. To address the non-differentiability of the $l_0$-norm, \citet{l0-regularization} propose a reparameterization of the neural network weights as the product of a weight and a stochastic gate variable sampled from a hard-concrete distribution. The parameters of the hard-concrete distribution can be optimized directly using the reparameterization trick, and the expected $l_0$-norm can be computed using the value of the cumulative distribution function of the random gate variable evaluated at zero.

\begin{table}[t]
\caption{\textbf{Constant hyperparameters for all Transformer experiments.} More details on the standard configuration for training the Transformer can be found in \citet{transformer}.}
\label{transformer-hparams}
\begin{tabular}{cc}
\hline
Hyperparameter & Value \\ \hline
dataset                     & translate\_wmt\_ende\_packed        \\
training iterations         & 500000                              \\
batch size                  & 2048 tokens                         \\
learning rate schedule      & standard transformer\_base          \\
optimizer                   & Adam                                \\
sparsity range              & 50\% - 98\%                         \\
beam search                 & beam size 4; length penalty 0.6     \\ \hline
\end{tabular}
\vskip -0.2in
\end{table}

\subsection{Random Pruning Baseline}

For our experiments, we also include a random sparsification procedure adapted from the magnitude pruning technique of \citet{to-prune-or-not}. Our random pruning technique uses the same sparsity schedule, but differs by selecting the weights to be pruned each step at random rather based on magnitude and does not allow pruned weights to reactivate. This technique is intended to represent a lower-bound of the accuracy-sparsity trade-off curve. 

\subsection{Experimental Framework}

For magnitude pruning, we used the TensorFlow model pruning library. We implemented variational dropout and $l_0$ regularization from scratch. For variational dropout, we verified our implementation by reproducing the results from the original paper. To verify our $l_0$ regularization implementation, we applied our weight-level code to Wide ResNet \cite{wide-resnet} trained on CIFAR-10 and replicated the training FLOPs reduction and accuracy results from the original publication. Verification results for variational dropout and $l_0$ regularization are included in Appendices \ref{appendix_vd_repro} and \ref{appendix_l0_repro}. For random pruning, we modified the TensorFlow model pruning library to randomly select weights as opposed to sorting them based on magnitude. 

For each model, we kept the number of training steps constant across all techniques and performed extensive hyper-parameter tuning. While magnitude pruning is relatively simple to apply to large models and achieves 
reasonably consistent performance across a wide range of hyperparameters, variational dropout and $l_0$-regularization are much less well understood. To our knowledge, we are the first to apply these techniques to models of this scale. To produce a fair comparison, we did not limit the amount of hyperparameter tuning we performed for each technique. In total, our results encompass over 4000 experiments.

\begin{figure}[t]
\begin{center}
\centerline{\includegraphics[width=\columnwidth]{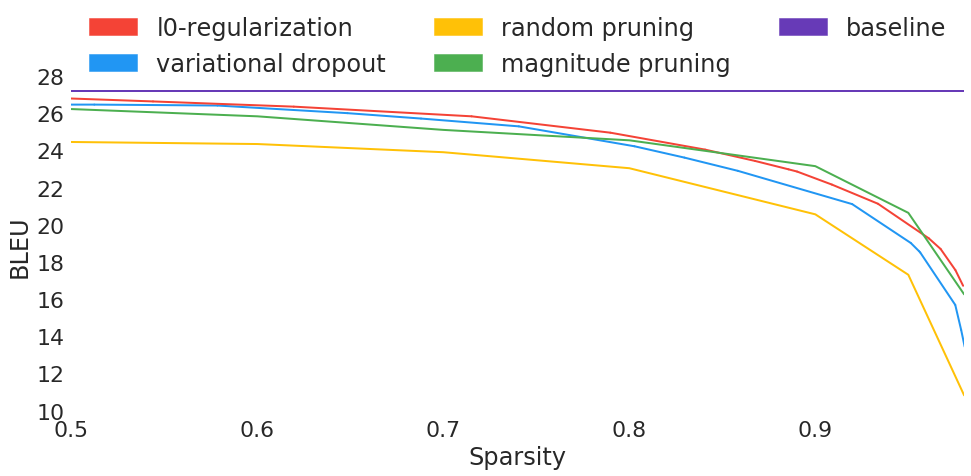}}
\centerline{\includegraphics[width=\columnwidth]{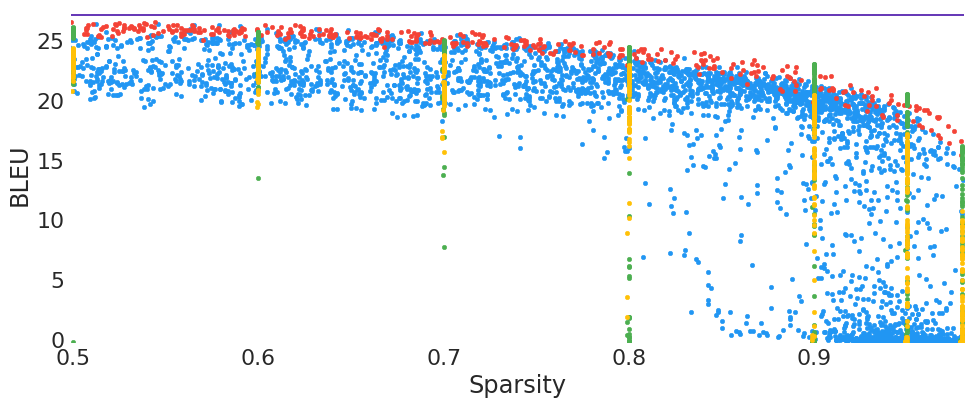}}
\caption{\textbf{Sparsity-BLEU trade-off curves for the Transformer.} Top: Pareto frontiers for each of the four sparsification techniques applied to the Transformer. Bottom: All experimental results with each technique. Despite the diversity of approaches, the relative performance of all three techniques is remarkably consistent. Magnitude pruning notably outperforms more complex techniques for high levels of sparsity.}
\label{sparse_transformer}
\end{center}
\vskip -0.42in
\end{figure}

\section{Sparse Neural Machine Translation}

We adapted the Transformer \cite{transformer} model for neural machine translation to use these four sparsification techniques, and trained the model on the WMT 2014 English-German dataset. We sparsified all fully-connected layers and embeddings, which make up 99.87\% of all of the parameters in the model (the other parameters coming from biases and layer normalization). The constant hyperparameters used for all experiments are listed in table \ref{transformer-hparams}. We followed the standard training procedure used by \citet{transformer}, but did not perform checkpoint averaging. This setup yielded a baseline BLEU score of 27.29 averaged across five runs.

We extensively tuned the remaining hyperparameters for each technique. Details on what hyperparameters we explored, and the results of what settings produced the best models can be found in Appendix \ref{transformer_hparams}.

\subsection{Sparse Transformer Results \& Analysis}

All results for the Transformer are plotted in figure \ref{sparse_transformer}. Despite the vast differences in these approaches, the relative performance of all three techniques is remarkably consistent. While $l_0$ regularization and variational dropout produce the top performing models in the low-to-mid sparsity range, magnitude pruning achieves the best results for highly sparse models. While all techniques were able to outperform the random pruning technique, randomly removing weights produces surprisingly reasonable results, which is perhaps indicative of the models ability to recover from damage during optimization.

\begin{figure}[t]
\vskip 0.2in \begin{center}
\centerline{\includegraphics[width=1\columnwidth]{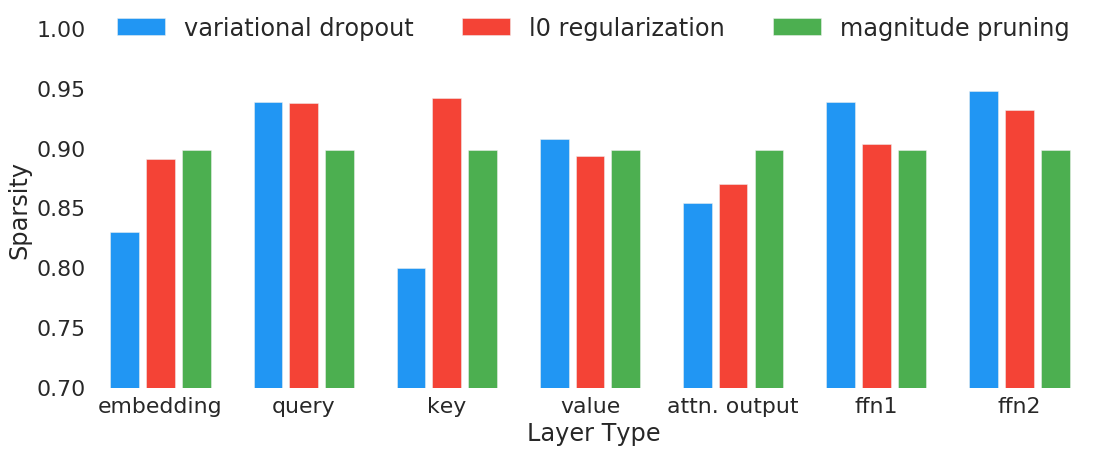}}
\caption{\textbf{Average sparsity in Transformer layers.} Distributions calculated on the top performing model at 90\% sparsity for each technique. $l_0$ regularization and variational dropout are able to learn non-uniform distributions of sparsity, while magnitude pruning induces user-specified sparsity distributions (in this case, uniform).}
\label{transformer_sparsity_distributions}
\end{center} \vskip -0.8in \end{figure}

What is particularly notable about the performance of magnitude pruning is that our experiments uniformly remove the same fraction of weights for each layer. This is in stark contrast to variational dropout and $l_0$ regularization, where the distribution of sparsity across the layers is learned through the training process. Previous work has shown that a non-uniform sparsity among different layers is key to achieving high compression rates \cite{automatic-model-compression}, and variational dropout and $l_0$ regularization should theoretically be able to leverage this feature to learn better distributions of weights for a given global sparsity. 

Figure \ref{transformer_sparsity_distributions} shows the distribution of sparsity across the different layer types in the Transformer for the top performing model at 90\% global sparsity for each technique. Both $l_0$ regularization and variational dropout learn to keep more parameters in the embedding, FFN layers, and the output transforms for the multi-head attention modules and induce more sparsity in the transforms for the query and value inputs to the attention modules. Despite this advantage, $l_0$ regularization and variational dropout did not significantly outperform magnitude pruning, even yielding inferior results at high sparsity levels.

It is also important to note that these results maintain a constant number of training steps across all techniques and that the Transformer variant with magnitude pruning trains 1.24x and 1.65x faster than $l_0$ regularization and variational dropout respectively. While the standard Transformer training scheme produces excellent results for machine translation, it has been shown that training the model for longer can improve its performance by as much as 2 BLEU \cite{scaling-nmt}. Thus, when compared for a fixed training cost magnitude pruning has a distinct advantage over these more complicated techniques.

\section{Sparse Image Classification}

To benchmark these four sparsity techniques on a large-scale computer vision task, we integrated each method into ResNet-50 and trained the model on the ImageNet large-scale image classification dataset. We sparsified all convolutional and fully-connected layers, which make up 99.79\% of all of the parameters in the model (the other parameters coming from biases and batch normalization).

\begin{table}[t]
\caption{\textbf{Constant hyperparameters for all RN50 experiments.}}
\resizebox{\columnwidth}{!}{%
\begin{tabular}{cc}
\hline
Hyperparameter & Value \\
\hline
\footnotesize dataset                     & \footnotesize ImageNet        \\
\footnotesize training iterations         & \footnotesize 128000                              \\
\footnotesize batch size                  & \footnotesize 1024 images                        \\
\footnotesize learning rate schedule      & \footnotesize standard        \\
\footnotesize optimizer                   & \footnotesize SGD with Momentum                                \\
\footnotesize sparsity range              & \footnotesize 50\% - 98\% \\ \hline
\end{tabular}%
}
\label{tab:rn50-hparams}
\vskip -0.1in
\end{table}

The hyperparameters we used for all experiments are listed in Table \ref{tab:rn50-hparams}. Each model was trained for 128000 iterations with a batch size of 1024 images, stochastic gradient descent with momentum, and the standard learning rate schedule (see Appendix \ref{resnet_lr}). This setup yielded a baseline top-1 accuracy of 76.69\% averaged across three runs. We trained each model with 8-way data parallelism across 8 accelerators. Due to the extra parameters and operations required for variational dropout, the model was unable to fit into device memory in this configuration. For all variational dropout experiments, we used a per-device batch size of 32 images and scaled the model over 32 accelerators.

\begin{figure}[t]
\begin{center}
\centerline{\includegraphics[width=1.05\columnwidth]{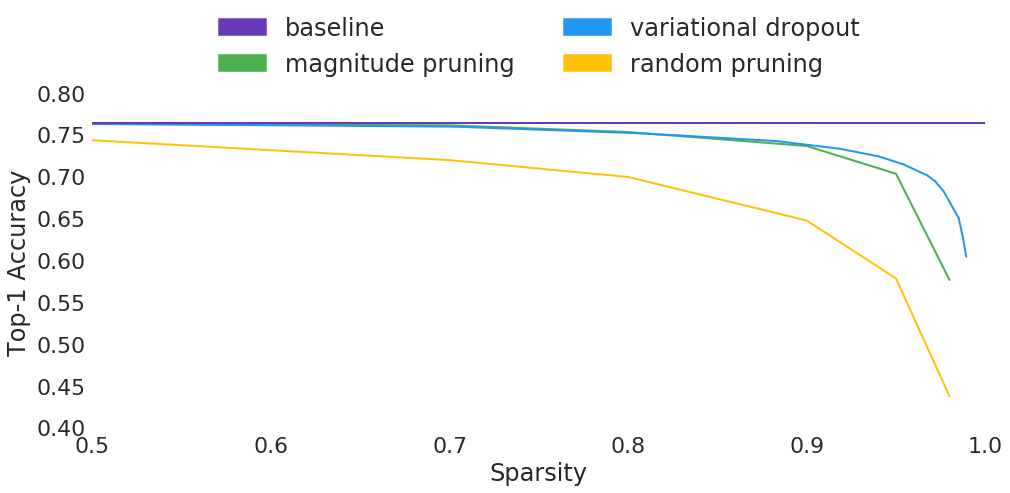}}
\centerline{\includegraphics[width=1.05\columnwidth]{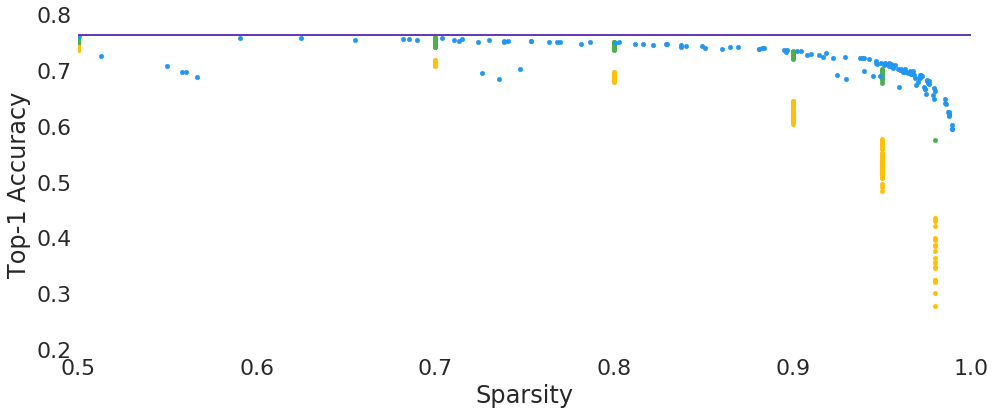}}
\caption{\textbf{Sparsity-accuracy trade-off curves for ResNet-50.} Top: Pareto frontiers for variational dropout, magnitude pruning, and random pruning applied to ResNet-50. Bottom: All experimental results with each technique. We observe large variation in performance for variational dropout and $l_0$ regularization between Transformer and ResNet-50. Magnitude pruning and variational dropout achieve comparable performance for most sparsity levels, with variational dropout achieving the best results for high sparsity levels.}
\label{sparse_rn50}
\end{center}
\vskip -0.4in
\end{figure}

\subsection{ResNet-50 Results \& Analysis}

Figure \ref{sparse_rn50} shows results for magnitude pruning, variational dropout, and random pruning applied to ResNet-50. Surprisingly, we were unable to produce sparse ResNet-50 models with $l_0$ regularization that did not significantly damage model quality. Across hundreds of experiments, our models were either able to achieve full test set performance with no sparsification, or sparsification with test set performance akin to random guessing. Details on all hyperparameter settings explored are included in Appendix \ref{resnet_hparams}.

This result is particularly surprising given the success of $l_0$ regularization on Transformer. One nuance of the $l_0$ regularization technique of \citet{l0-regularization} is that the model can have varying sparsity levels between the training and test-time versions of the model. At training time, a parameter with a dropout rate of 10\% will be zero 10\% of the time when sampled from the hard-concrete distribution. However, under the test-time parameter estimator, this weight will be non-zero.\footnote{The fraction of time a parameter is set to zero during training depends on other factors, e.g. the $\beta$ parameter of the hard-concrete distribution. However, this point is generally true that the training and test-time sparsities are not necessarily equivalent, and that there exists some dropout rate threshold below which a weight that is sometimes zero during training will be non-zero at test-time.}. \citet{l0-regularization} reported results applying $l_0$ regularization to a wide residual network (WRN) \cite{wide-resnet} on the CIFAR-10 dataset, and noted that they observed small accuracy loss at as low as 8\% reduction in the number of parameters during training. Applying our weight-level $l_0$ regularization implementation to WRN produces a model with comparable training time sparsity, but with no sparsity in the test-time parameters. For models that achieve test-time sparsity, we observe significant accuracy degradation on CIFAR-10. This result is consistent with our observation for $l_0$ regularization applied to ResNet-50 on ImageNet.

The variation in performance for variational dropout and $l_0$ regularization between Transformer and ResNet-50 is striking. While achieving a good accuracy-sparsity trade-off, variational dropout consistently ranked behind $l_0$ regularization on Transformer, and was bested by magnitude pruning for sparsity levels of 80\% and up. However, on ResNet-50 we observe that variational dropout consistently produces models on-par or better than magnitude pruning, and that $l_0$ regularization is not able to produce sparse models at all. Variational dropout achieved particularly notable results in the high sparsity range, maintaining a top-1 accuracy over 70\% with less than 4\% of the parameters of a standard ResNet-50.

\begin{figure}[t]
\begin{center}
\centerline{\includegraphics[width=1.05\columnwidth]{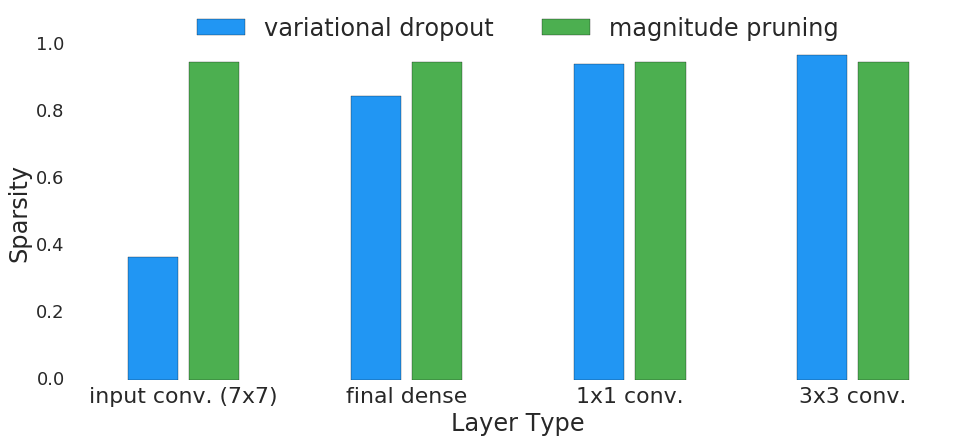}}
\caption{\textbf{Average sparsity in ResNet-50 layers.} Distributions calculated on the top performing model at 95\% sparsity for each technique. Variational dropout is able to learn non-uniform distributions of sparsity, decreasing sparsity in the input and output layers that are known to be disproportionately important to model quality.}
\label{rn50_sparsity_distributions}
\end{center}
\end{figure}

The distribution of sparsity across different layer types in the best variational dropout and magnitude pruning models at 95\% sparsity are plotted in figure \ref{rn50_sparsity_distributions}. While we kept sparsity constant across all layers for magnitude and random pruning, variational dropout significantly reduces the amount of sparsity induced in the first and last layers of the model.

It has been observed that the first and last layers are often disproportionately important to model quality \cite{lwac, deep-rewiring}. In the case of ResNet-50, the first convolution comprises only .037\% of all the parameters in the model. At 98\% sparsity the first layer has only 188 non-zero parameters, for an average of less than 3 parameters per output feature map. With magnitude pruning uniformly sparsifying each layer, it is surprising that it is able to achieve any test set performance at all with so few parameters in the input convolution.

While variational dropout is able to learn to distribute sparsity non-uniformly across the layers, it comes at a significant increase in resource requirements. For ResNet-50 trained with variational dropout we observed a greater than 2x increase in memory consumption. When scaled across 32 accelerators, ResNet-50 trained with variational dropout completed training in 9.75 hours, compared to ResNet-50 with magnitude pruning finishing in 12.50 hours on only 8 accelerators. Scaled to a 4096 batch size and 32 accelerators, ResNet-50 with magnitude pruning can complete the same number of epochs in just 3.15 hours.

\subsection{Pushing the Limits of Magnitude Pruning}

Given that a uniform distribution of sparsity is suboptimal, and the significantly smaller resource requirements for applying magnitude pruning to ResNet-50 it is natural to wonder how well magnitude pruning could perform if we were to distribute the non-zero weights more carefully and increase training time. 

\begin{figure}[t]
\begin{center}
\centerline{\includegraphics[width=\columnwidth]{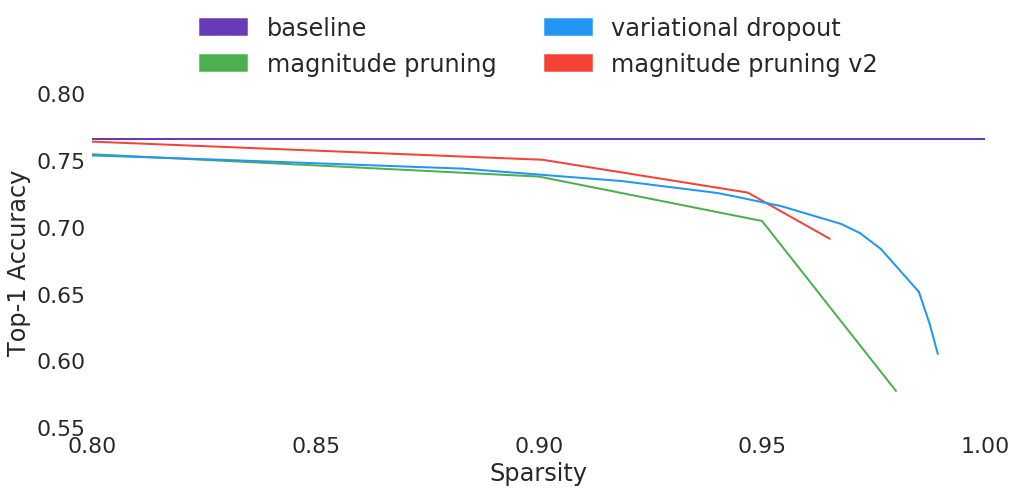}}
\caption{\textbf{Sparsity-accuracy trade-off curves for ResNet-50 with modified sparsification scheme.} Altering the distribution of sparsity across the layers and increasing training time yield significant improvement for magnitude pruning.}
\label{rn50-mp-supertune}
\end{center}
\vskip -0.3in
\end{figure}

To understand the limits of the magnitude pruning heuristic, we modify our ResNet-50 training setup to leave the first convolutional layer fully dense, and only prune the final fully-connected layer to 80\% sparsity. This heuristic is reasonable for ResNet-50, as the first layer makes up a small fraction of the total parameters in the model and the final layer makes up only .03\% of the total FLOPs. While tuning the magnitude pruning ResNet-50 models, we observed that the best models always started and ended pruning during the third learning rate phase, before the second learning rate drop. To take advantage of this, we increase the number of training steps by 1.5x by extending this learning rate region. Results for ResNet-50 trained with this scheme are plotted in figure \ref{rn50-mp-supertune}.

With these modifications, magnitude pruning outperforms variational dropout at all but the highest sparsity levels while still using less resources. However, variational dropout's performance in the high sparsity range is particularly notable. With very low amounts of non-zero weights, we find it likely that the models performance on the test set is closely tied to precise allocation of weights across the different layers, and that variational dropout's ability to learn this distribution enables it to better maintain accuracy at high sparsity levels. This result indicates that efficient sparsification techniques that are able to learn the distribution of sparsity across layers are a promising direction for future work.

Its also worth noting that these changes produced models at 80\% sparsity with top-1 accuracy of 76.52\%, only .17\% off our baseline ResNet-50 accuracy and .41\% better than the results reported by \citet{automatic-model-compression}, without the extra complexity and computational requirements of their reinforcement learning approach. This represents a new state-of-the-art sparsity-accuracy trade-off for ResNet-50 trained on ImageNet.

\section{Sparsification as Architecture Search}

While sparsity is traditionally thought of as a model compression technique, two independent studies have recently suggested that the value of sparsification in neural networks is misunderstood, and that 
once a sparse topology is learned it can be trained from scratch to the full performance achieved when sparsification was performed jointly with optimization.

\citet{lottery-ticket-hypothesis} posited that over-parameterized neural networks contain small, trainable subsets of weights, deemed "winning lottery tickets". They suggest that sparsity inducing techniques are methods for finding these sparse topologies, and that once found the sparse architectures can be trained from scratch with \textit{the same weight initialization that was used when the sparse architecture was learned}. They demonstrated that this property holds across different convolutional neural networks and multi-layer perceptrons trained on the MNIST and CIFAR-10 datasets.

\citet{rethinking-pruning} similarly demonstrated this phenomenon for a number of activation sparsity techniques on convolutional neural networks, as well as for weight level sparsity learned with magnitude pruning. However, they demonstrate this result using a random initialization during re-training.

The implications of being able to train sparse architectures from scratch once they are learned are large: once a sparse topology is learned, it can be saved and shared as with any other neural network architecture. Re-training then can be done fully sparse, taking advantage of sparse linear algebra to greatly accelerate time-to-solution. However, the combination of these two studies does not clearly establish how this potential is to be realized.

Beyond the question of whether or not the original random weight initialization is needed, both studies only explore convolutional neural networks (and small multi-layer perceptrons in the case of \citet{lottery-ticket-hypothesis}). The majority of experiments in both studies also limited their analyses to the MNIST, CIFAR-10, and CIFAR-100 datasets. While these are standard benchmarks for deep learning models, they are not indicative of the complexity of real-world tasks where model compression is most useful. \citet{rethinking-pruning} do explore convolutional architectures on the ImageNet datasets, but only at two relatively low sparsity levels (30\% and 60\%). They also note that weight level sparsity on ImageNet is the only case where they are unable to reproduce the full accuracy of the pruned model.

To clarify the questions surrounding the idea of sparsification as a form of neural architecture search, we repeat the experiments of \citet{lottery-ticket-hypothesis} and \citet{rethinking-pruning} on ResNet-50 and Transformer. For each model, we explore the full range of sparsity levels (50\% - 98\%) and compare to our well-tuned models from the previous sections.

\begin{figure}[t]
\begin{center}
\centerline{\includegraphics[width=1.05\columnwidth]{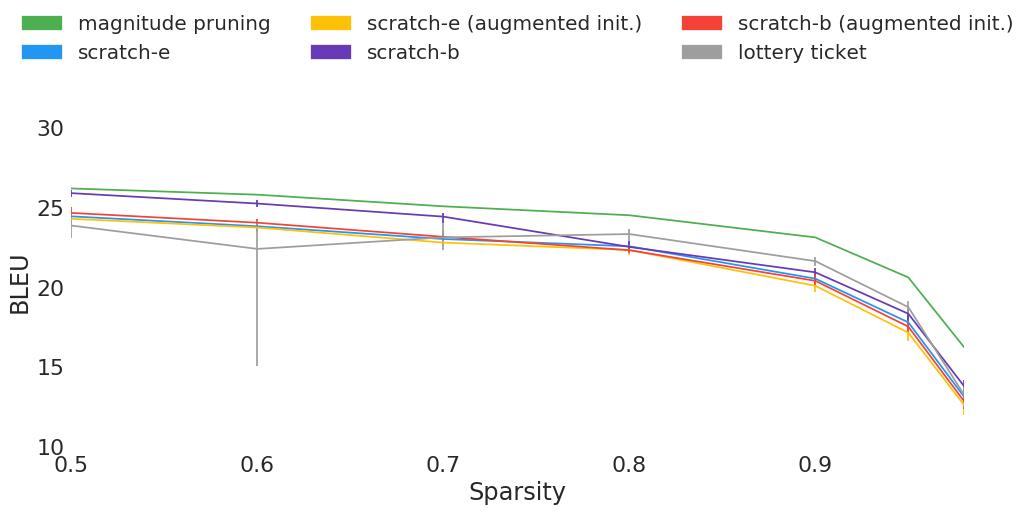}}
\centerline{\includegraphics[width=1.05\columnwidth]{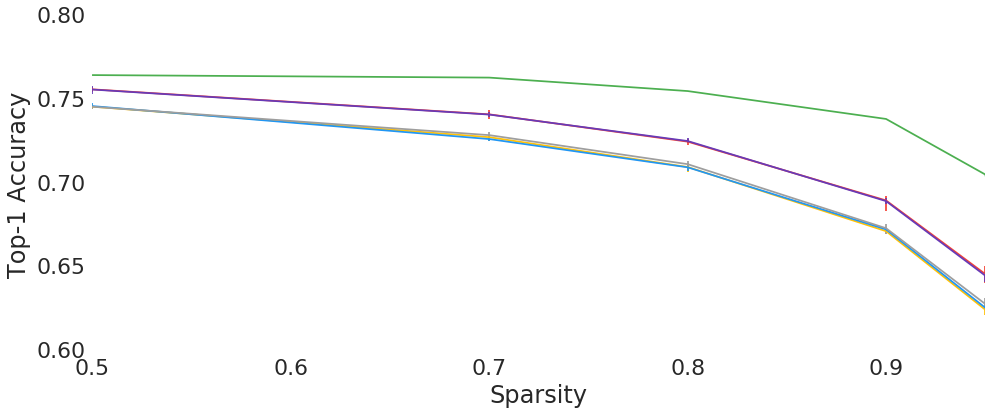}}
\caption{\textbf{Scratch and lottery ticket experiments with magnitude pruning.} Top: results with Transformer. Bottom: Results with ResNet-50. Across all experiments, training from scratch using a learned sparse architecture is unable to re-produce the performance of models trained with sparsification as part of the optimization process.}
\label{scratch_lotto}
\end{center}
\vskip -0.4in
\end{figure}

\subsection{Experimental Framework}

The experiments of \citet{rethinking-pruning} encompass taking the final learned weight mask from a magnitude pruning model, randomly re-initializing the weights, and training the model with the normal training procedure (i.e., learning rate, number of iterations, etc.). To account for the presence of sparsity at the start of training, they scale the variance of the initial weight distribution by the number of non-zeros in the matrix. They additionally train a variant where they increase the number of training steps (up to a factor of 2x) such that the re-trained model uses approximately the same number of FLOPs during training as model trained with sparsification as part of the optimization process. They refer to these two experiments as "scratch-e" and "scratch-b" respectively.

\citet{lottery-ticket-hypothesis} follow a similar procedure, but use the same weight initialization that was used when the sparse weight mask was learned and do not perform the longer training time variant.

For our experiments, we repeat the scratch-e, scratch-b and lottery ticket experiments with magnitude pruning on Transformer and ResNet-50. For scratch-e and scratch-b, we also train variants that do not alter the initial weight distribution. For the Transformer, we re-trained five replicas of the best magnitude pruning hyperparameter settings at each sparsity level and save the weight initialization and final sparse weight mask. For each of the five learned weight masks, we train five identical replicas for the scratch-e, scratch-b, scratch-e with augmented initialization, scratch-b with augmented initialization, and the lottery ticket experiments. For ResNet-50, we followed the same procedure with three re-trained models and three replicas at each sparsity level for each of the five experiments. Figure \ref{scratch_lotto} plots the averages and min/max of all experiments at each sparsity level\footnote{Two of the 175 Transformer experiments failed to train from scratch at all and produced BLEU scores less than 1.0. We omit these outliers in figure \ref{scratch_lotto}}.

\subsection{Scratch and Lottery Ticket Results \& Analysis}

Across all of our experiments, we observed that training from scratch using a learned sparse architecture is not able to match the performance of the same model trained with sparsification as part of the optimization process.

Across both models, we observed that doubling the number of training steps did improve the quality of the results for the scratch experiments, but was not sufficient to match the test set performance of the magnitude pruning baseline. As sparsity increased, we observed that the deviation between the models trained with magnitude pruning and those trained from scratch increased. For both models, we did not observe a benefit from using the augmented weight initialization for the scratch experiments.

For ResNet-50, we experimented with four different learning rates schemes for the scratch-b experiments. We found that scaling each learning rate region to double the number of epochs produced the best results by a wide margin. These results are plotted in figure \ref{scratch_lotto}. Results for the ResNet-50 scratch-b experiments with the other learning rate variants are included with our release of hyperparameter tuning results.

For the lottery ticket experiments, we were not able to replicate the phenomenon observed by \citet{lottery-ticket-hypothesis}. The key difference between our experiments is the complexity of the tasks and scale of the models, and it seems likely that this is the main factor contributing to our inability to train these architecture from scratch.

For the scratch experiments, our results are consistent with the negative result observed by \cite{rethinking-pruning} for ImageNet and ResNet-50 with unstructured weight pruning. By replicating the scratch experiments at the full range of sparsity levels, we observe that the quality of the models degrades relative to the magnitude pruning baseline as sparsity increases. For unstructured weight sparsity, it seems likely that the phenomenon observed by \citet{rethinking-pruning} was produced by a combination of low sparsity levels and small-to-medium sized tasks. We'd like to emphasize that this result is only for unstructured weight sparsity, and that prior work \citet{rethinking-pruning} provides strong evidence that activation pruning behaves differently.

\section{Limitations of This Study}
\label{limitations}

\textbf{Hyperparameter exploration.} For all techniques and models, we carefully hand-tuned hyperparameters and performed extensive sweeps encompassing thousands of experiments over manually identified ranges of values. However, the number of possible settings vastly outnumbers the set of values that can be practically explored, and we cannot eliminate the possibility that some techniques significantly outperform others under settings we did not try. 

\textbf{Neural architectures and datasets.} Transformer and ResNet-50 were chosen as benchmark tasks to represent a cross section of large-scale deep learning tasks with diverse architectures. We can't exclude the possibility that some techniques achieve consistently high performance across other architectures. More models and tasks should be thoroughly explored in future work.

\section{Conclusion}

In this work, we performed an extensive evaluation of three state-of-the-art sparsification techniques on two large-scale learning tasks. Notwithstanding the limitations discussed in section \ref{limitations}, we demonstrated that complex techniques shown to yield state-of-the-art compression on small datasets perform inconsistently, and that simple heuristics can achieve comparable or better results on a reduced computational budget. Based on insights from our experiments, we achieve a new state-of-the-art sparsity-accuracy trade-off for ResNet-50 with only magnitude pruning and highlight promising directions for research in sparsity inducing techniques.

Additionally, we provide strong counterexamples to two recently proposed theories that models learned through pruning techniques can be trained from scratch to the same test set performance of a model learned with sparsification as part of the optimization process. Our results highlight the need for large-scale benchmarks in sparsification and model compression. As such, we open-source our code, checkpoints, and results of all hyperparameter configurations to establish rigorous baselines for future work.

\section*{Acknowledgements}

We would like to thank Benjamin Caine, Jonathan Frankle, Raphael Gontijo Lopes, Sam Greydanus, and Keren Gu for helpful discussions and feedback on drafts of this paper.

\bibliography{main}
\bibliographystyle{icml2019}

\twocolumn[
\icmltitle{The State of Sparsity in Deep Neural Networks: Appendix}
]
\appendix
\icmltitlerunning{The State of Sparsity in Deep Neural Networks: Appendix}

\section{Overview of Sparsity Inducing Techniques}
\label{appendix_a}

Here we provide a more detailed review of the three sparsity techniques we benchmarked.

\subsection{Magnitude Pruning}

Magnitude-based weight pruning schemes use the magnitude of each weight as a proxy for its importance to model quality, and remove the least important weights according to some sparsification schedule over the course of training. Many variants have been proposed \cite{memory-bounded-convnet, lwac, dynamic-network-surgery, to-prune-or-not}, with the key differences lying in when weights are removed, whether weights should be sorted to remove a precise proportion or thresholded based on a fixed or decaying value, and whether or not weights that have been pruned still receive gradient updates and have the potential to return after being pruned.

\citet{lwac} use iterative magnitude pruning and re-training to progressively sparsify a model. The target model is first trained to convergence, after which a portion of weights are removed and the model is re-trained with these weights fixed to zero. This process is repeated until the target sparsity is achieved. \citet{dynamic-network-surgery} improve on this approach by allowing masked weights to still receive gradient updates, enabling the network to recover from incorrect pruning decisions during optimization. They achieve higher compression rates and interleave pruning steps with gradient update steps to avoid expensive re-training. \citet{to-prune-or-not} similarly allow gradient updates to masked weights, and make use of a gradual sparsification schedule with sorting-based weight thresholding to maintain accuracy while achieving a user specified level of sparsification. 

It’s worth noting that magnitude pruning can easily be adapted to induce block or activation level sparsity by removing groups of weights based on their p-norm, average, max, or other statistics. Variants have also been proposed that maintain a constant level of sparsity during optimization to enable accelerated training \cite{sparse-evolutionary-training}.

\subsection{Variational Dropout}

Consider the setting of a dataset $\mathcal{D}$ of $N$ i.i.d. samples $(\mathbf{x}, \mathbf{y})$ and a standard classification problem where the goal is to learn the parameters $\mathbf{w}$ of the conditional probability $p(\mathbf{y} | \mathbf{x}, \mathbf{w})$. Bayesian inference combines some initial belief over the parameters $\mathbf{w}$ in the form of a prior distribution $p(\mathbf{w})$ with observed data $\mathcal{D}$ into an updated belief over the parameters in the form of the posterior distribution $p(\mathbf{w} | \mathcal{D})$. In practice, computing the true posterior using Bayes' rule is computationally intractable and good approximations are needed. In variational inference, we optimize the parameters $\phi$ of some parameterized model $q_{\phi}(\mathbf{w})$ such that $q_{\phi}(\mathbf{w})$ is a close approximation to the true posterior distribution $p(\mathbf{w} | \mathcal{D})$ as measured by the Kullback-Leibler divergence between the two distributions. The divergence of our approximate posterior from the true posterior is minimized in practice by maximizing the variational lower-bound 

$$\mathcal{L}(\phi) = -D_{KL}(q_{\phi}(\mathbf{w}) || p(\mathbf{w})) + L_{\mathcal{D}}(\phi)$$

where $L_{\mathcal{D}}(\phi) = \sum\limits_{(\mathbf{x}, \mathbf{y}) \in \mathcal{D}} \mathbf{E}_{q_{\phi}(\mathbf{w})}[\textrm{log} \ p(\mathbf{y} | \mathbf{x}, \mathbf{w})]$

Using the Stochastic Gradient Variational Bayes (SGVB) \cite{variational-dropout-local-reparameterization} algorithm to optimize this bound, $L_{\mathcal{D}}(\phi)$ reduces to the standard cross-entropy loss, and the KL divergence between our approximate posterior and prior over the parameters serves as a regularizer that enforces our initial belief about the parameters $\mathbf{w}$.

In the standard formulation of variational dropout, we assume the weights are drawn from a fully-factorized Gaussian approximate posterior.

$$w_{ij} \sim q_{\phi}(w_{ij}) = \mathcal{N}(\theta_{ij}, \alpha_{ij} \theta_{ij}^2)$$

Where $\theta$ and $\alpha$ are neural network parameters. For each training step, we sample weights from this distribution and use the \textit{reparameterization trick} \cite{autoencoding-variational-bayes, stochastic-backpropagation} to differentiate the loss w.r.t. the parameters through the sampling operation. Given the weights are normally distributed, the distribution of the activations $\mathbf{B}$ after a linear operation like matrix multiplication or convolution is also Gaussian and can be calculated in closed form \footnote{We ignore correlation in the activations, as is done by \citet{variational-dropout}}.

$$q_{\phi}(b_{mj} | \mathbf{A}) \sim \mathcal{N}(\gamma_{mj}, \delta_{mj})$$

with $\gamma_{mj} = \sum\limits_{i=1}^K a_{mi} \theta_{ij}$ and $\delta_{mj} = \sum\limits_{i=1}^K a_{mi}^2 \alpha_{ij} \theta_{ij}^2$ and where $a_{mi} \in \mathbf{A}$ are the inputs to the layer. Thus, rather than sample weights, we can directly sample the activations at each layer. This step is known as the \textit{local reparameterization trick}, and was shown by \citet{variational-dropout-local-reparameterization} to reduce the variance of the gradients relative to the standard formulation in which a single set of sampled weights must be shared for all samples in the input batch for efficiency. \citet{variational-dropout} showed that the variance of the gradients could be further reduced by using an \textit{additive noise reparameterization}, where we define a new parameter 

$$\sigma_{ij}^2 = \alpha_{ij}*\theta_{ij}^2$$ 

Under this parameterization, we directly optimize the mean and variance of the neural network parameters.

Under the assumption of a log-uniform prior on the weights $\mathbf{w}$, the KL divergence component of our objective function $D_{KL}(q_{\phi}(w_{ij}) || p(w_{ij}))$ can be accurately approximated \cite{variational-dropout}:

\begin{gather*}
D_{KL}(q_{\phi}(w_{ij}) || p(w_{ij})) \approx \\
k_1 \sigma(k_2 + k_3 \ \textrm{log} \ \alpha_{ij}) - 0.5 \ \textrm{log}(1 + \alpha_{ij}^{-1} + -k_1) \\
k_1 = 0.63576 \quad k_2 = 1.87320 \quad k_3 = 1.48695
\end{gather*}

After training a model with variational dropout, the weights with the highest $\alpha$ values can be removed. For all their experiments, \citet{variational-dropout} removed weights with $\textrm{log} \ \alpha$ larger than 3.0, which corresponds to a dropout rate greater than 95\%. Although they demonstrated good results, it is likely that the optimal $\alpha$ threshold varies across different models and even different hyperparameter settings of the same model. We address this question in our experiments.

\subsection{$l_0$ Regularization}

To optimize the $l_0$-norm, we reparameterize the model weights $\theta$ as the product of a weight and a random variable drawn from the hard-concrete distribution.

\begin{gather*}
\theta_j = \tilde{\theta}_jz_j \\
\textrm{where} \ z_j \sim \textrm{min}(1, \textrm{max}(0, \overline{s})), \ \overline{s} = s(\zeta - \gamma) + \gamma \\
s = \textrm{sigmoid}((\textrm{log} \ u - \textrm{log}(1 - u) + \textrm{log} \ \alpha) / \beta) \\
\textrm{and} \ u \sim \mathcal{U}(0, 1) 
\end{gather*}

In this formulation, the $\alpha$ parameter that controls the position of the hard-concrete distribution (and thus the probability that $z_j$ is zero) is optimized with gradient descent. $\beta$, $\gamma$, and $\zeta$ are fixed parameters that control the shape of the hard-concrete distribution. $\beta$ controls the curvature or \textit{temperature} of the hard-concrete probability density function, and $\gamma$ and $\zeta$ stretch the distribution s.t. $z_j$ takes value 0 or 1 with non-zero probability. 

On each training iteration, $z_j$ is sampled from this distribution and multiplied with the standard neural network weights. The expected $l_0$-norm $\mathcal{L}_C$ can then be calculated using the cumulative distribution function of the hard-concrete distribution and optimized directly with stochastic gradient descent.

$$\mathcal{L}_C = \sum\limits_{j=1}^{|\theta|}(1 - Q_{\overline{s}_j}(0 | \phi)) = \sum\limits_{j=1}^{|\theta|} \textrm{sigmoid}(\textrm{log} \ \alpha_j - \beta \ \textrm{log} \ \frac{-\gamma}{\zeta})$$

At test-time, \citet{l0-regularization} use the following estimate for the model parameters.

\begin{gather*}
\boldsymbol{\theta^*} = \boldsymbol{\tilde{\theta}^*} \odot \hat{\mathbf{z}} \\
\hat{\mathbf{z}} = \textrm{min}(\mathbf{1}, \textrm{max}(\mathbf{0}, \textrm{sigmoid}(\textrm{log} \ \boldsymbol{\alpha})(\zeta - \gamma) + \gamma))
\end{gather*}

Interestingly, \citet{l0-regularization} showed that their objective function under the $l_0$ penalty is a special case of a variational lower-bound over the parameters of the network under a spike and slab \cite{spike-and-slab} prior.

\section{Variational Dropout Implementation Verification}
\label{appendix_vd_repro}
To verify our implementation of variational dropout, we applied it to LeNet-300-100 and LeNet-5-Caffe on MNIST and compared our results to the original paper \cite{variational-dropout}. We matched our hyperparameters to those used in the code released with the paper\footnote{https://github.com/ars-ashuha/variational-dropout-sparsifies-dnn}. All results are listed in table \ref{vd_repro}

\begin{table}[ht]
\centering
\caption{\textbf{Variational Dropout MNIST Reproduction Results.}}
\resizebox{\columnwidth}{!}{%
\label{vd_repro}
\begin{tabular}{llll}
\hline
\textbf{Network} & \textbf{Experiment} & \textbf{Sparsity (\%)} & \textbf{Accuracy (\%)} \\
\hline
\multirow{4}{*}{LeNet-300-100} & original \cite{variational-dropout}               & 98.57                  & 98.08                  \\
                               & ours (log $\alpha$ = 3.0) & 97.52                  & 98.42                  \\
                               & ours (log $\alpha$ = 2.0) & 98.50                  & 98.40                  \\
                               & ours (log $\alpha$ = 0.1) & 99.10                  & 98.13                  \\ \hline
\multirow{3}{*}{LeNet-5-Caffe} & original \cite{variational-dropout}               & 99.60                  & 99.25                  \\
                               & ours (log $\alpha$ = 3.0) & 99.29                  & 99.26                  \\
                               & ours (log $\alpha$ = 2.0) & 99.50                  & 99.25 \\ \hline
\end{tabular}%
}
\end{table}

Our baseline LeNet-300-100 model achieved test set accuracy of 98.42\%, slightly higher than the baseline of 98.36\% reported in \cite{variational-dropout}. Applying our variational dropout implementation to LeNet-300-100 with these hyperparameters produced a model with 97.52\% global sparsity and 98.42\% test accuracy. The original paper produced a model with 98.57\% global sparsity, and 98.08\% test accuracy. While our model achieves .34\% higher tests accuracy with 1\% lower sparsity, we believe the discrepancy is mainly due to difference in our software packages: the authors of \cite{variational-dropout} used Theano and Lasagne for their experiments, while we use TensorFlow. 

Given our model achieves highest accuracy, we can decrease the $\textrm{log} \ \alpha$ threshold to trade accuracy for more sparsity. With a $\textrm{log} \ \alpha$ threshold of 2.0, our model achieves 98.5\% global sparsity with a test set accuracy of 98.40\%.  With a $\textrm{log} \ \alpha$ threshold of 0.1, our model achieves 99.1\% global sparsity with 98.13\% test set accuracy, exceeding the sparsity and accuracy of the originally published results.

On LeNet-5-Caffe, our implementation achieved a global sparsity of 99.29\% with a test set accuracy of 99.26\%, versus the originaly published results of 99.6\% sparsity with 99.25\% accuracy. Lowering the $\textrm{log} \ \alpha$ threshold to 2.0, our model achieves 99.5\% sparsity with 99.25\% test accuracy.

\section{$l_0$ Regularization Implementation Verification}
\label{appendix_l0_repro}
The original $l_0$ regularization paper uses a modified version of the proposed technique for inducing group sparsity in models, so our weight-level implementation is not directly comparable. However, to verify our implementation we trained a Wide ResNet (WRN) \cite{wide-resnet} on CIFAR-10 and compared results to those reported in the original publication for group sparsity.

\begin{figure}[t]
\begin{center}
\centerline{\includegraphics[width=\columnwidth]{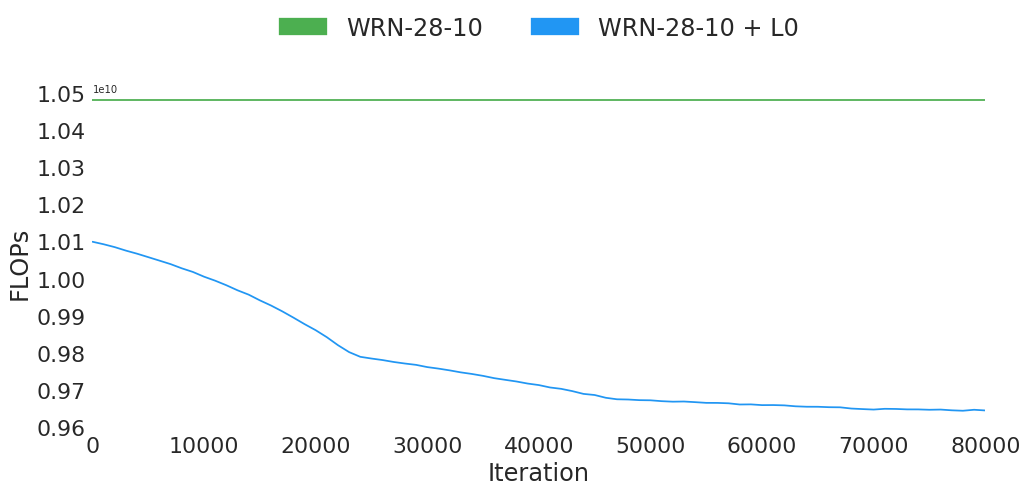}}
\caption{\textbf{Forward pass FLOPs for WRN-28-10 trained with $l_0$ regularization.} Our implementation achieves FLOPs reductions comparable to those reported in \citet{l0-regularization}.}
\label{wrn-l0}
\end{center}
\vskip -.3in
\end{figure}

As done by \citet{l0-regularization}, we apply $l_0$ to the first convolutional layer in the residual blocks (i.e., where dropout would normally be used). We use the weight decay formulation for the re-parameterized weights, and scale the weight decay coefficient to maintain the same initial length scale of the parameters. We use the same batch size of 128 samples and the same initial log $\alpha$, and train our model on a single GPU.

Our baseline WRN-28-10 implementation trained on CIFAR-10 achieved a test set accuracy of 95.45\%. Using our $l_0$ regularization implementation and a $l_0$-norm weight of .0003, we trained a model that achieved 95.34\% accuracy on the test set while achieving a consistent training-time FLOPs reduction comparable to that reported by \citet{l0-regularization}. Floating-point operations (FLOPs) required to compute the forward over the course of training WRN-28-10 with $l_0$ are plotted in figure \ref{wrn-l0}.

During our re-implementation of the WRN experiments from \citet{l0-regularization}, we identified errors in the original publications FLOP calculations that caused the number of floating-point operations in WRN-28-10 to be miscalculated. We've contacted the authors, and hope to resolve this issue to clarify their performance results.

\section{Sparse Transformer Experiments}
\label{transformer_hparams}

\subsection{Magnitude Pruning Details}

For our magnitude pruning experiments, we tuned four key hyperparameters: the starting iteration of the sparsification process, the ending iteration of the sparsification process, the frequency of pruning steps, and the combination of other regularizers (dropout and label smoothing) used during training. We trained models with 7 different target sparsities: 50\%, 60\%, 70\%, 80\%, 90\%, 95\%, and 98\%. At each of these sparsity levels, we tried pruning frequencies of 1000 and 10000 steps. During preliminary experiments we identified that the best settings for the training step to stop pruning at were typically closer to the end of training. Based on this insight, we explored every possible combination of start and end points for the sparsity schedule in increments of 100000 steps with an ending step of 300000 or greater.

By default, the Transformer uses dropout with a dropout rate of 10\% on the input to the encoder, decoder, and before each layer and performs label smoothing with a smoothing parameter of .1. We found that decreasing these other regularizers produced higher quality models in the mid to high sparsity range. For each hyperparameter combination, we tried three different regularization settings: standard label smoothing and dropout, label smoothing only, and no regularization.

\subsection{Variational Dropout Details}

For the Transformer trained with variational dropout, we extensively tuned the coefficient for the KL divergence component of the objective function to find models that achieved high accuracy with sparsity levels in the target range. We found that KL divergence weights in the range $[\frac{.1}{N}, \frac{1}{N}]$, where $N$ is the number of samples in the training set, produced models in our target sparsity range.

\cite{variational-dropout} noted difficulty training some models from scratch with variational dropout, as large portions of the model adopt high dropout rates early in training before the model can learn a useful representation from the data. To address this issue, they use a gradual ramp-up of the KL divergence weight, linearly increasing the regularizer coefficient until it reaches the desired value. 

For our experiments, we explored using a constant regularizer weight, linearly increasing the regularizer weight, and also increasing the regularizer weight following the cubic sparsity function used with magnitude pruning. For the linear and cubic weight schedules, we tried each combination of possible start and end points in increments of 100000 steps. For each hyperparameter combination, we also tried the three different combinations of dropout and label smoothing as with magnitude pruning. For each trained model, we evaluated the model with 11 $\textrm{log} \ \alpha$ thresholds in the range $[0, 5]$. For all experiments, we initialized all $\textrm{log} \ \sigma^2$ parameters to the constant value $-10$.

\subsection{$l_0$ Regularization Details}

For Transformers trained with $l_0$ regularization, we similarly tuned the coefficient for the $l_0$-norm in the objective function. We observed that much higher magnitude regularization coefficients were needed to produce models with the same sparsity levels relative to variational dropout. We found that $l_0$-norm weights in the range $[\frac{1}{N}, \frac{10}{N}]$ produced models in our target sparsity range.

For all experiments, we used the default settings for the paramters of the hard-concrete distribution: $\beta = 2/3$, $\gamma = -0.1$, and $\zeta = 1.1$. We initialized the $\textrm{log} \ \alpha$ parameters to $2.197$, corresponding to a 10\% dropout rate.

For each hyperparameter setting, we explored the three regularizer coefficient schedules used with variational dropout and each of the three combinations of dropout and label smoothing.

\subsection{Random Pruning Details}

We identified in preliminary experiments that random pruning typically produces the best results by starting and ending pruning early and allowing the model to finish the rest of the training steps with the final sparse weight mask. For our experiments, we explored all hyperparameter combinations that we explored with magnitude pruning, and also included start/end pruning step combinations with an end step of less than 300000.

\section{Sparse ResNet-50}
\label{resnet_hparams}

\subsection{Learning Rate}
\label{resnet_lr}

For all experiments, the we used the learning rate scheme used by the official TensorFlow ResNet-50 implementation\footnote{https://bit.ly/2Wd2Lk0}. With our batch size of 1024, this includes a linear ramp-up for 5 epochs to a learning rate of .4 followed by learning rate drops by a factor of 0.1 at epochs 30, 60, and 80.

\subsection{Magnitude Pruning Details}

For magnitude pruning on ResNet-50, we trained models with a target sparsity of 50\%, 70\%, 80\%, 90\%, 95\%, and 98\%. At each sparsity level, we tried starting pruning at steps 8k, 20k, and 40k. For each potential starting point, we tried ending pruning at steps 68k, 76k, and 100k. For every hyperparameter setting, we tried pruning frequencies of 2k, 4k, and 8k steps and explored training with and without label smoothing. During preliminary experiments, we observed that removing weight decay from the model consistently caused significant decreases in test accuracy. Thus, for all hyperparameter combinations, we left weight decay on with the standard coefficient.

For a target sparsity of 98\%, we observed that very few hyperparameter combinations were able to complete training without failing due to numerical issues. Out of all the hyperparameter configurations we tried, only a single model was able to complete training without erroring from the presence of NaNs. As explained in the main text, at high sparsity levels the first layer of the model has very few non-zero parameters, leading to instability during training and low test set performance. Pruned ResNet-50 models with the first layer left dense did not exhibit these issues.

\subsection{Variational Dropout Details}

For variational dropout applied to ResNet-50, we explored the same combinations of start and end points for the kl-divergence weight ramp up as we did for the start and end points of magnitude pruning. For all transformer experiments, we did not observe a significant gain from using a cubic kl-divergence weight ramp-up schedule and thus only explored the linear ramp-up for ResNet-50. For each combination of start and end points for the kl-divergence weight, we explored 9 different coefficients for the kl-divergence loss term: .01 / N, .03 / N, .05 / N, .1 / N, .3 / N, .5 / N, 1 / N, 10 / N, and 100 / N. 

Contrary to our experience with Transformer, we found ResNet-50 with variational dropout to be highly sensitive to the initialization for the log $\sigma^2$ parameters. With the standard setting of -10, we couldn't match the baseline accuracy, and with an initialization of -20 our models achieved good test performance but no sparsity. After some experimentation, we were able to produce good results with an initialization of -15.

While with Transformer we saw a reasonable amount of variance in test set performance and sparsity with the same model evaluated at different log $\alpha$ thresholds, we did not observe the same phenomenon for ResNet-50. Across a range of log $\alpha$ values, we saw consistent accuracy and nearly identical sparsity levels. For all of the results reported in the main text, we used a log $\alpha$ threshold of 0.5, which we found to produce slightly better results than the standard threshold of 3.0.

\subsection{$l_0$ Regularization Details}

For $l_0$ regularization, we explored four different initial log $\alpha$ values corresponding to dropout rates of 1\%, 5\%, 10\%, and 30\%. For each dropout rate, we extenively tuned the $l_0$-norm weight to produce models in the desired sparsity range. After identifying the proper range of $l_0$-norm coefficients, we ran experiments with 20 different coefficients in that range. For each combination of these hyperparameters, we tried all four combinations of other regularizers: standard weight decay and label smoothing, only weight decay, only label smoothing, and no regularization. For weight decay, we used the formulation for the reparameterized weights provided in the original paper, and followed their approach of scaling the weight decay coefficient based on the initial dropout rate to maintain a constant length-scale between the $l_0$ regularized model and the standard model.

Across all of these experiments, we were unable to produce ResNet models that achieved a test set performance better than random guessing. For all experiments, we observed that training proceeded reasonably normally until the $l_0$-norm loss began to drop, at which point the model incurred severe accuracy loss. We include the results of all hyperparameter combinations in our data release.

Additionally, we tried a number of tweaks to the learning process to improve the results to no avail. We explored training the model for twice the number of epochs, training with much higher initial dropout rates, modifying the $\beta$ parameter for the hard-concrete distribution, and a modified test-time parameter estimator. 

\subsection{Random Pruning Details}

For random pruning on ResNet-50, we shifted the set of possible start and end points for pruning earlier in training relative to those we explored for magnitude pruning. At each of the sparsity levels tried with magnitude pruning, we tried starting pruning at step 0, 8k, and 20k. For each potential starting point, we tried ending pruning at steps 40k, 68k, and 76k. For every hyperparameter setting, we tried pruning frequencies of 2k, 4k, and 8k and explored training with and without label smoothing.

\subsection{Scratch-B Learning Rate Variants}

For the scratch-b \cite{rethinking-pruning} experiments with ResNet-50, we explored four different learning rate schemes for the extended training time (2x the default number of epochs).

The first learning rate scheme we explored was uniformly scaling each of the five learning rate regions to last for double the number of epochs. This setup produced the best results by a wide margin. We report these results in the main text.

The second learning rate scheme was to keep the standard learning rate, and maintain the final learning rate for the extra training steps as is common when fine-tuning deep neural networks. The third learning rate scheme was to maintain the standard learning rate, and continually drop the learning rate by a factor of 0.1 every 30 epochs. The last scheme we explored was to skip the learning rate warm-up, and drop the learning rate by 0.1 every 30 epochs. This learning rate scheme is closest to the one used by \citet{rethinking-pruning}. We found that this scheme underperformed relative to the scaled learning rate scheme with our training setup.

Results for all learning rate schemes are included with the released hyperparameter tuning data.

\end{document}